\documentclass[a4paper]{article}

\usepackage{INTERSPEECH2018}

\usepackage{epsfig,amssymb,amsmath}
\usepackage{multirow}
\usepackage{booktabs}
\usepackage{url}
\usepackage[pdftex]{color}

\usepackage{tikz}
\usetikzlibrary{bayesnet}
\usetikzlibrary{arrows,shapes,backgrounds,positioning,fit}

\title{Fast variational Bayes for heavy-tailed PLDA applied to i-vectors and x-vectors}
\name{Anna Silnova$^1$, Niko Br\"ummer$^2$, Daniel Garcia-Romero$^3$, David Snyder$^3$, and Luk\'a\v s Burget$^1$}
\address{
  $^1$Brno University of Technology, Czech Republic\\
	$^2$Nuance Communications, South Africa\\
  $^3$Johns Hopkins HLTCOE, USA}
\email{\{isilnova,burget\}@fit.vutbr.za, niko.brummer@gmail.com, dgromero@jhu.edu}

\DeclareMathOperator*{\argmax}{argmax}

\newcommand{\cprm}{$C^{\rm Prm}_{\rm min}$}
\newcommand{\mindcf}{$\rm minDCF_{\rm 0.01}$}

\def\R{\mathbb{R}}

\def\Lset{\mathcal{L}}
\def\Rset{\mathcal{R}}

\def\zvec{\mathbf{z}}
\def\avec{\mathbf{a}}
\def\rvec{\mathbf{r}}

\def\nulvec{\boldsymbol{0}}

\def\lambdavec{\boldsymbol{\lambda}}

\def\Imat{\mathbf{I}}
\def\Bmat{\mathbf{B}}
\def\Gmat{\mathbf{G}}
\def\Wmat{\mathbf{W}}

\def\Fmat{\mathbf{F}}

\def\expvb#1#2{\left\langle#1\right\rangle_{#2}}
\def\expv#1#2{\bigl\langle#1\bigr\rangle_{#2}}

\def\expp#1{\bigl\langle#1\bigr\rangle}

\def\ND{\mathcal{N}}
\def\TD{\mathcal{T}}
\def\GD{\mathcal{G}}

\def\const{\text{const}}

\def\barB{\bar\Bmat}

\begin{document}

\maketitle
\begin{abstract}
The standard state-of-the-art backend for text-independent speaker recognizers that use i-vectors or x-vectors, is Gaussian PLDA (G-PLDA), assisted by a Gaussianization step involving length normalization. G-PLDA can be trained with both generative or discriminative methods. It has long been known that heavy-tailed PLDA (HT-PLDA), applied without length normalization, gives similar accuracy, but at considerable extra computational cost. We have recently introduced a fast scoring algorithm for a discriminatively trained HT-PLDA backend. This paper extends that work by introducing a fast, variational Bayes, generative training algorithm. We compare old and new backends, with and without length-normalization, with i-vectors and x-vectors, on  SRE'10, SRE'16 and SITW.

\end{abstract}
\noindent\textbf{Index Terms}: speaker recognition, variational Bayes, heavy-tailed PLDA

\section{Introduction}
We extend our previous work in~\cite{Brummer_Odyssey18}, where we did discriminative training of a heavy-tailed PLDA model (HT-PLDA), applied to i-vectors. In this paper, we explore instead a generative training solution and we apply it to both i-vectors~\cite{ivec} and x-vectors~\cite{DSIS17}. The advantage of the generative training is that it is orders of magnitude faster than the discriminative one.

In~\cite{HTPLDA} HT-PLDA was shown to be a better model of i-vectors than Gaussian PLDA (G-PLDA), but the computational cost was considerable. Subsequently,~\cite{Dani_length_norm} showed that the i-vectors could instead be Gaussianized via a simple length normalization procedure. This matched the accuracy of HT-PLDA, with negligible extra computational cost and established G-PLDA with length-normalization as the standard back-end for scoring text-independent i-vector speaker recognizers. Recently, the same scoring recipe was applied also to x-vectors~\cite{DSIS17,DS_ICASSP18,SRI_xvector}.

In this paper, we revisit HT-PLDA, using a slightly simplified model for which we present fast training and scoring algorithms, with speed comparable to G-PLDA. The HT-PLDA model can be applied without length normalization. We demonstrate accuracy gains on some data sets, including EER=2.7\% on SITW and EER=3.2\% on SRE'16 Cantonese. Since we have effectively removed the computational impediment, we encourage other researchers to experiment with this backend as an alternative. We make open source code available to facilitate such experiments.


\section{HT-PLDA model}
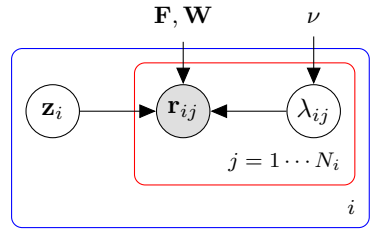
\begin{figure}[htb!]
\centerline{
\begin{tikzpicture}
\node[obs] (r) {$\rvec_{ij}$};
\node[const, outer sep = 1pt, above left = 4 pt of r] (ldummy) {};
\node[const, outer sep = 3pt, above right = 4 pt of r] (rdummy) {};
\node[latent, left = of r] (z) {$\zvec_i$};
\node[latent, right = of r] (lambda) {$\lambda_{ij}$};
\node[const, above = 2 em of lambda, inner sep = 5pt] (Gamma) {$\nu$};
\node[const, inner sep = 5pt] at(Gamma-|r)  (FW) {$\Fmat,\Wmat$};
\plate[draw=red,inner sep = 5pt] {red} {(ldummy)(r)(lambda)} {$j=1\cdots N_i$};
\plate[draw=blue,inner sep = 5pt] {} {(z)(red)} {$i$};
\edge {z}{r};
\edge {lambda}{r};
\edge {Gamma}{lambda};
\edge {FW}{r};
\end{tikzpicture}
} 
\caption{Heavy-tailed PLDA}
\label{fig:model}
\end{figure}

The generative HT-PLDA model is shown in graphical model notation~\cite{PRML} in figure~\ref{fig:model} and is defined as follows. For every speaker, $i$, let all of the available observations of that speaker ($N_i$ of them) be denoted as $\Rset_i=\{\rvec_{ij}\}_{j=1}^{N_i}$, where the $\rvec_{ij}\in\R^D$ are i-vectors, or x-vectors, of dimension $D$. For every speaker, a hidden speaker identity variable, $\zvec_i\in\R^d$, is drawn i.i.d.\ from the standard $d$-dimensional normal distribution. We require $d\ll D$. The heavy-tailed behaviour is obtained by drawing for every observation a hidden \emph{precision scaling factor}, $\lambda_{ij}>0$, from a gamma distribution, $\GD(\alpha,\beta)$ parametrized by $\alpha=\beta=\frac\nu2>0$. The parameter $\nu$ is known as the \emph{degrees of freedom}~\cite{HTPLDA,PRML}. Finally, given the hidden variables, the observations are drawn i.i.d.\ from the multivariate normal:
\begin{align}
P(\rvec_{ij}\mid\zvec_i,\lambda_{ij}) &= \ND\bigl(\rvec_{ij}\mid \Fmat\zvec_i,(\lambda_{ij}\Wmat)^{-1}\bigr)  
\end{align}
where $\Fmat$ is the $D$-by-$d$, \emph{factor loading matrix} and where $\Wmat$ is a $D$-by-$D$ positive definite \emph{precision matrix}. The model parameters are $\nu,\Fmat,\Wmat$. This model is a simplification of Kenny's HT-PLDA model~\cite{HTPLDA}, which also had heavy-tailed speaker identity variables.

This model does not allow closed-form scoring or training. Either the hidden scaling factors, or the hidden speaker identity variables can be integrated out in closed form, but not both. This means that we have to find approximations for both scoring and training. We make use of a new approximation, the \emph{Gaussian likelihood approximation}, as recently published in~\cite{Brummer_Odyssey18}. In that paper, the approximation was used for both scoring and discriminative training. In this paper, we apply the same approximation also for generative training.

\subsection{The Gaussian likelihood approximation}
\def\zhat{\hat\zvec}
Both scoring and training recipes can be built around the \emph{likelihood for the hidden speaker identity variable}, given the observation. Marginalization over the hidden variable, $\lambda_{ij}$, gives a multivariate t-distribution for the observed vector~\cite{PRML,meta_embeddings,Brummer_Odyssey18}:
\begin{align}
P(\rvec_{ij}\mid\zvec_i) &= \TD(\rvec_j \mid \Fmat\zvec_i, \Wmat,\nu)
\end{align}
This is a t-distribution for $\rvec_{ij}$, but to use this as likelihood for $\zvec_i$, we need to view it as function of $\zvec_i$. Provided that $D>d$ and $\Fmat'\Wmat\Fmat$ is invertible, it is shown in~\cite{meta_embeddings} that this function is proportional to another t-distribution, with \emph{increased} degrees of freedom, $\nu'=\nu+D-d$:
\begin{align}
P(\rvec_{ij}\mid\zvec_i) &\propto \TD(\zvec\mid \zhat_{ij},\Bmat_{ij},\nu')
\end{align} 
where 
\begin{align}
\zhat_{ij}&=\Bmat_{ij}^{-1}\avec_{ij} & \Bmat_{ij} &= b_{ij} \Bmat_0, \\
\label{eq:Banda}
\avec_{ij} &= b_{ij} \Fmat'\Wmat\rvec_{ij}, &
b_{ij} &= \frac{\nu+D-d}{\nu+\rvec'_{ij}\Gmat\rvec_{ij}}, \\
\Bmat_0 &= \Fmat'\Wmat\Fmat, &
\Gmat &= \Wmat-\Wmat\Fmat\Bmat_0^{-1}\Fmat'\Wmat
\end{align}
In a typical PLDA model, we have $d\in [100,200]$ and $D\in [400,600]$, so that $\nu'=\nu+D-d$ is large, making the likelihood practically Gaussian. We therefore approximate the speaker identity likelihood as:
\begin{align}
\label{eq:SGME_approx}
P(\rvec_{ij}\mid\zvec) &\approx \exp\bigl[\avec'_{ij}\zvec-\frac12\zvec'\Bmat_{ij}\zvec\bigr]
\propto \ND(\zvec\mid \zhat_{ij},\Bmat_{ij}^{-1}) 
\end{align}
Notice that for the heavy-tailed case, with small $\nu$, the likelihood precisions are variable, while in the Gaussian limit, at $\nu\to\infty$, we have $b_{ij}=1$ and constant precisions. The precision variability is driven by $\rvec'_{ij}\Gmat\rvec_{ij}$, where $\Gmat$ is a projection operator onto the orthogonal complement of the speaker subspace\footnote{$\Gmat\Fmat=\nulvec$}---with the inner product defined by the positive definite precision matrix~\cite{ob_proj}.

\subsection{Scoring}
Since the precisions of all likelihoods extracted by this model differ only by a scale factor, $b_{ij}$, they can be jointly diagonalized, to give fast scoring of speaker verification trials as explained in more detail in~\cite{Brummer_Odyssey18,meta_embeddings}.  

\subsection{Mean-field VB training}
\label{sec:vb}
\def\barB{\bar\Bmat}
\def\zbar{\bar\zvec}
The hidden variables associated with speaker $i$ are $\zvec_i$ and $\lambdavec_i=\{\lambda_{ij}\}_{j=1}^{N_i}$. These hidden variables are \emph{dependent} in the true joint posterior, $P(\zvec_i,\lambdavec_i\mid\Rset_i)$ and this posterior does not have a closed form. We follow~\cite{HTPLDA} and use mean-field variational Bayes (VB)~\cite{PRML} that makes use of an approximate, factorized posterior of the form:
\begin{align}
Q_i(\lambdavec,\zvec) &= Q_i(\lambdavec)Q_i(\zvec) \approx P(\zvec,\lambdavec\mid\Rset_i)
\end{align}
Given this factorization, the \emph{VB lower bound} is formed as:
\begin{align}
\label{eq:VBL}
\Lset &= \sum_i \expvb{\log\frac{P(\Rset_i,\zvec,\lambdavec\mid\theta)}{Q_i(\zvec)Q_i(\lambdavec)}}{Q_i(\zvec)Q_i(\lambdavec)}
\end{align}
where the model parameters are $\theta=(\Fmat,\Wmat,\nu)$; and where $\Lset\le P(\Rset\mid\theta)$, the true marginal likelihood. Training is done by maximizing $\Lset$ w.r.t.\ both $\theta$ and the $Q$-factors, to give an approximation to $\argmax_{\theta}P(\Rset\mid\theta)$. 

In the traditional mean-field VB recipe, $\Lset$ is optimized iteratively, doing partial maximizations w.r.t.\ $\theta$, the $Q_i(\lambdavec)$  and the $Q_i(\zvec)$ in round-robin fashion. The $Q$-factor optimizations are variational, rather than parametric. The \emph{forms} of the optimal $Q$-factors can be derived in closed form: multivariate Gaussian for $Q_i(\zvec)$; and product of independent gamma distributions for $Q_i(\lambdavec)$. But for every $i$, the parameters of these distributions have to be iteratively computed. This makes the traditional recipe slow. To get a fast alternative, we choose a closed form for the one $Q$-factor:
\begin{align}
\label{eq:Qlambda}
Q_i(\lambdavec_i) &= \prod_{j=1}^{N_i} \GD\bigl(\lambda_{ij}\mid \frac{\nu+D-d}2,\frac{\nu+\rvec'_{ij}\Gmat\rvec_{ij}}2\bigr)
\end{align}
This is \emph{different} from the optimal mean-field factor, but this choice is made such that the expected values, $\expp{\lambda_{ij}}$ equal the $b_{ij}$ in ~\eqref{eq:Banda}. Given~\eqref{eq:Qlambda}, we can now apply the standard mean-field solution~\cite{PRML}:
\begin{align}
\log Q_i(\zvec) &= \expv{\log P(\Rset_i,\zvec\mid\lambdavec)}{Q_i(\lambdavec)} + \const
\end{align}
 to also find the other $Q$-factor in closed form:     
\begin{align}
\label{eq:Qz}
Q_i(\zvec) &\propto P(\zvec) \prod_{j=1}^{N_i} \ND(\zvec\mid\zhat_{ij},\Bmat_{ij}^{-1}) 
\propto \ND(\zvec\mid\zbar_i,\barB_i^{-1})
\end{align}
where $P(\zvec)=\ND(\zvec\mid\nulvec,\Imat)$ and
\begin{align}
\zbar_i &= \barB_i^{-1} \sum_{j=1}^{N_i} \avec_{ij}, &
\barB_i &= \Imat+\sum_{j=1}^{N_i}\Bmat_{ij}
\end{align}
This solution agrees with the Gaussian likelihood approximation in the following sense. Notice that the true posterior for $\zvec_i$ can be expressed as:
\begin{align}
P(\zvec\mid\Rset_i)&\propto P(\zvec)\prod_j P(\rvec_{ij}\mid\zvec)
\end{align} 
If we replace the above t-distribution likelihoods, $P(\rvec_{ij}\mid\zvec)$, with the Gaussian approximations in~\eqref{eq:SGME_approx}, then we also arrive at~\eqref{eq:Qz}. 

Our learning algorithm proceeds as follows. Fix $\nu$ to some chosen value and randomly initialize $\Fmat,\Wmat$. Then iterate:
\begin{description}
		\item[E-step:] Assign the $Q_i(\lambdavec)$ and $Q_i(\zvec)$ using~\eqref{eq:Qlambda} and~\eqref{eq:Qz}.
		\item[M-step:] Maximize~\eqref{eq:VBL} w.r.t.\ $\Fmat,\Wmat$.
\end{description}
All of the above steps have closed forms and the resulting optimization algorithm proceeds very similarly to the usual EM-algorithm for training Gaussian PLDA~\cite{SPLDA}. The only difference is that the zero, first and second-order stats in the new algorithm are weighted by the scaling factors, $b_{ij}$. As in~\cite{SPLDA}, we augment the M-step with minimum divergence~\cite{mindiv} on the hidden variable prior $P(\zvec)$. We also do minimum divergence on the hidden scale factors, as explained in \emph{Example 2.6: Multivariate t-distribution with known degrees of freedom}, in~\cite{EM4T}. The minimum divergence augmentations lead to much faster convergence and a well-calibrated end-result. Again, the posterior precisions, $\barB_i$, are mutually diagonalizable, requiring but a single eigenanalysis of $\Bmat_0$ per iteration.

An open-source implementation of the training and scoring algorithms for this model is available at \url{github.com/bsxfan/meta-embeddings/tree/master/code/Niko/matlab/clean/VB4HTPLDA}.

\section{Experiments}
\subsection{HT-PLDA for i-vectors}
\subsubsection{Experimental setup}
We continue our experiments with HT-PLDA modeling of i-vectors started in \cite{Brummer_Odyssey18}. Here, we keep the same experimental setup except for a few minor differences mentioned below.  

As before, spectral features are 60-dimensional MFCCs with short-term mean and variance normalization applied over a 3 second sliding window. The UBM is gender independent and has 2048 diagonal components. The i-vectors are of dimension $D=600$. We applied global mean normalization to these i-vectors (because our HT-PLDA model does not have a mean parameter). The G-PLDA backend was applied to i-vectors both with and without length normalization (LN). All HT-PLDA backends were applied to i-vectors without LN.

UBM, i-vector extractor and both Gaussian and heavy-tailed PLDAs are trained on the PRISM dataset~\cite{PRISM}, containing Fisher parts 1 and 2, Switchboard 2, 3 and Switchboard cellphone phases. Also, NIST SRE 2004--2008 (also known as MIXER collections) are added to the training. In total, the set contains  approximately 100K utterances coming from 16241 speakers. We used 8000 randomly selected files for UBM training and the full set to train the i-vector extractor. When training PLDA models, we filter out all speakers having less than 6 utterances, resulting in just 3429 speakers and 73306 training utterances.

We evaluate performance on the female part of NIST SRE'10, condition 5, which consists of English telephone data \cite{NISTSRE10_evalplan}. Additionally, we report the results on the NIST SRE'16 evaluation set (both males and females). We report the results on the whole evaluation set as well as on two language subsets, Cantonese and Tagalog. As evaluation metrics, we use the equal error rate (EER, in \%) as well as the average minimum detection cost function for two operating points of interest in the NIST SRE'16 \cite{NIST_SRE_2016} ($C^{\rm Prm}_{\rm min}$).

\subsubsection{Experiments and results}
Our i-vector experiments, comparing traditional G-PLDA with discriminatively and generatively trained HT-PLDA are shown in table~\ref{tbl:results_ivecs}. The speaker subspace dimensionality is $d=200$. The first two lines show the G-PLDA baseline, with and without length normalization. As expected, length normalization helps to shape the distribution of i-vectors to better fit Gaussian assumptions made by G-PLDA. Consequently, the performance of G-PLDA is significantly worse when no length normalization is applied. 

The third and fourth lines repeat our experiments from~\cite{Brummer_Odyssey18}. Line 3 shows HT-PLDA with $\nu=2$ and with $\Fmat,\Wmat$ simply initialized from G-PLDA.\footnote{HT-PLDA with $\nu\to\infty$ is equivalent to G-PLDA.} Line 4 shows the same model after additional discriminative training with binary cross-entropy (BXE). In both cases no length normalization was applied. In the absence of LN, introducing the heavy-tailed mechanism at test time (line 3) is able to significantly improve the performance (compared to line 2), even without further training. Discriminatively trained HT-PLDA, without length norm (line 4) does best and surpasses the performance of G-PLDA with length normalization (line 1).

The last two lines of table~\ref{tbl:results_ivecs} present results for generative VB training of HT-PLDA as described in section~\ref{sec:vb}. In line 5, training was done with $\nu\to\infty$ and testing with $\nu=2$. In line 6, both training and testing had $\nu=2$. Ideally, lines 5 and 3 should be identical, but due to details of the respective EM and VB algorithms, small differences remain, possibly because the algorithms were stopped before fully converged. Although the VB trained HT-PLDA variants (lines 5, 6) do better than G-PLDA baseline without LN (line 2), it does not manage to improve the performance of G-PLDA with LN (line 1), nor of the discriminatively trained HT-PLDA (line 4). 

\begin{table*}
\caption{\label{tbl:results_ivecs} Comparison of accuracies on SRE 2010 and 2016 of Gaussian PLDA with length normalization and without it, versus heavy-tailed PLDA (without length normalization). The performance metrics are \cprm, and \rm{EER}(\%)}
\vspace{3mm}
\centerline{
\begin{tabular}{l c c c c c c c c c c c c}
\toprule
\multirow{2}{*}{System}   &  \multicolumn{2}{c}{SRE10 c05,f} & & \multicolumn{2}{c}{SRE16, all} & & \multicolumn{2}{c}{SRE16, Cantonese}& &\multicolumn{2}{c}{SRE16, Tagalog}\\
\cmidrule{2-3}
\cmidrule{5-6}
\cmidrule{8-9}
\cmidrule{11-12}
  & \cprm  & EER & & \cprm  & EER & & \cprm  & EER & & \cprm  & EER\\
  \midrule
G-PLDA, LN &  0.26 &  2.5 & &  0.97 & 16.5 & & \textbf{0.68} & 9.7 & & 0.99 & 21.0\\
G-PLDA, no LN &  0.33 &  4.0 & &  0.97 & 17.8 & & 0.69 & 11.5 & & 0.98 & 21.3 \\
\midrule
HT-PLDA $\nu=2$, initialized from G-PLDA &  0.30 &  2.9 & &  0.96 & 16.7 & & 0.68 & 10.0 & & 0.98 & 21.1 \\ 
HT-PLDA $\nu=2$, trained with BXE & \textbf{0.21} &  \textbf{2.1} & &  \textbf{0.90} & \textbf{15.1} & & 0.74 & \textbf{9.3} & & \textbf{0.97} & \textbf{20.2} \\
\midrule
HT-PLDA, train $\nu=\infty$, test $\nu=2$ &  0.30 &  2.7 & &  0.97 & 16.7 & & 0.68 & 10.0 & & 0.98 & 21.2\\
HT-PLDA, train $\nu=2$, test $\nu=2$ & 0.31 &  3.2 & &  0.97 & 16.9 & & 0.69 & 10.4 & & 0.99 & 21.3\\
\bottomrule
\end{tabular}}
\end{table*}

\subsection{HT-PLDA for x-vectors}

\subsubsection{x-vector extractor}

The x-vector system is a modified version of the DNN in \cite{DS_ICASSP18}. The features are 23 dimensional MFCCs with a frame-length of 25ms,
mean-normalized over a sliding window of up to 3 seconds. An energy SAD is used to filter out nonspeech frames. The first few layers of the x-vector extractor operate on sequences of frames. They are a hierarchy of convolutional layers (only convolving in time) that provide a long temporal context (23 frames, 11 to each side of the center frame) with reduced complexity. Their outputs are processed by fully connected layers and followed by a statistics pooling layer that aggregates across time by computing the mean and standard deviation. This process aggregates information so that subsequent layers operate on the entire segment. The mean and standard deviation are concatenated together
and propagated through segment-level layers and finally the softmax output layer. The nonlinearities are all rectified linear units (ReLUs).

The DNN is trained to classify the $N$ speakers in the training data.
A training example consists of a chunk of speech features (about 3 seconds average),
and the corresponding speaker label.
After training, x-vectors (512 dimension) are extracted from an affine layer 2 levels above the pooling layer.

The software framework has been made available in the Kaldi toolkit.
An example recipe is in the main branch of Kaldi at \url{https://github.com/kaldi-asr/kaldi/tree/master/egs/sre16/v2} and a pretrained x-vector system can be downloaded from \url{http://kaldi-asr.org/models.html}.

\subsubsection{Experimental setup}

The DNN training data consists of both telephone and microphone speech (mostly English). All wideband audio is downsampled to 8kHz. We pooled data from Switchboard, Fisher, Mixer (SRE 2004-2010), and VoxCeleb\footnote{We removed the 60 speakers that overlap with SITW since we evaluate on it.} \cite{Nagrani17} datasets yielding approximately 175K recordings from 15K speakers. Additionally, the recordings were augmented (using noise, reverb, and music) to produce 450K examples. From this augmented set, 15K chunks of 2 to 4 seconds were extracted for each speaker to form minibatches (64 chunks). We sampled equally for each speaker (i.e., balanced the training data per speaker) and trained for 3 epochs.

The G-PLDA and HT-PLDA classifiers are trained on a subset of the augmented data (we removed Switchboard and Fisher data) comprising 7K speakers and 230K recordings. For all experiments, we use a speaker subspace of dimension $d=150$. To explore the effects of LN on x-vectors we present results with and without it. More precisely, although LN comprises multiple steps (center, whitening, and projection onto unit-sphere) we use the notation ``no LN'' to refer to the lack of projection. We always center and whiten the data. Finally, the scores are normalized using adaptive symmetric score normalization (ass-norm)~\cite{sturim2005snorm}.

We report results on SITW core-core condition
\cite{mclaren2016} and the Cantonese subset of NIST SRE'16~\cite{NIST_SRE_2016} to characterize the system behavior under microphone and telephone recording conditions. Each of these sets provides development data that we use for centering the evaluation data and computing ass-norm. The PLDA training set is always centered to its own mean and used to estimate the whitening transform. Note that this transformation does not have any impact if no projection is applied to the x-vectors. Additionally, for the SRE'16 set, we also show results applying unsupervised domain adaptation~\cite{unsup_adapt} of the PLDA parameters.

\subsubsection{Results}
The x-vector results are presented in tables~\ref{tbl:results_xvecs_sitw} and~\ref{tbl:results_xvecs_sre16}. To the best of our knowledge, these are the best numbers published on both tasks. Moreover, the HT-PLDA classifier with no LN outperforms G-PLDA (even with domain adaptation for SRE'16). It is interesting to note that LN is detrimental to the HT-PLDA performance. Recall that the precision scaling factors $b_{ij}$ in~\eqref{eq:Banda} are determined by $\rvec'_{ij}\Gmat\rvec_{ij}$, the energy of the x-vectors in the complement of the speaker subspace. This results in scaling factors that get smaller as the energy of the x-vectors outside of the speaker subspace grows (which is consistent with the phenomenon that our model is trying to capture). Therefore, projecting the x-vector onto the unit sphere interferes with this process and the results indicate that it is detrimental. The G-PLDA classifier seems suboptimal for these tasks, but still benefits from LN. This is more noticeable for the SRE'16 results than for SITW where LN does not seem to have much effect. This is an indication that x-vectors behave differently than i-vectors in this regard and requires further investigation. Finally, unsupervised domain adaptation using parameter interpolation works quite well for both G-PLDA and the HT-PLDA model.

\begin{table}
\caption{\label{tbl:results_xvecs_sitw} G-PLDA vs HT-PLDA on eval part of SITW core-core using x-vectors.} 
\vspace{3mm}
\centerline{
\begin{tabular}{l c c }
\toprule
System  & \mindcf & EER \\
 \midrule
G-PLDA, LN &  0.34 & 3.3  \\
G-PLDA, no LN &  0.34 & 3.4\\
\midrule
HT-PLDA, $\nu=2$, LN& 0.34& 3.4 \\
HT-PLDA, $\nu=2$, no LN& \textbf{0.33} & \textbf{2.7} \\
\bottomrule
\end{tabular}}
\end{table}

\begin{table}
\caption{\label{tbl:results_xvecs_sre16} G-PLDA vs HT-PLDA (with and without adaptation) on SRE 2016 Cantonese using x-vectors. The performance metrics are balanced \cprm, as computed by NIST scoring tool,  \mindcf and \rm{EER}(\%)} 
\vspace{3mm}
\centerline{
\begin{tabular}{l c c c}
\toprule
System  & \cprm (bal.) & \mindcf & EER \\
 \midrule
G-PLDA, LN & 0.30 &  0.31 &  4.5 \\
G-PLDA, no LN & 0.33 &  0.32 &  4.7 \\
HT-PLDA, $\nu=2$, LN & 0.31& 0.31 &  4.5  \\
HT-PLDA, $\nu=2$, no LN & \textbf{0.30}& \textbf{0.30} &  \textbf{3.8}  \\
\midrule
+ unsupervised adaptation \\
\midrule
G-PLDA, LN & 0.27 &  0.27 &  3.9  \\
G-PLDA, no LN  & 0.29 &  0.28 &  4.3 \\
HT-PLDA, $\nu=2$, LN & 0.27 & 0.28 &  4.2  \\
HT-PLDA, $\nu=2$, no LN & \textbf{0.25 }& \textbf{0.26} &  \textbf{3.2}  \\
\bottomrule
\end{tabular}}
\end{table}

\section{Discussion}
In this paper and in our previous work~\cite{Brummer_Odyssey18}, we revisit heavy-tailed PLDA and re-engineer it to provide a computationally attractive alternative to the existing state of the art given by Gaussian PLDA with length normalization. Our experiments show benefits to HT-PLDA on both i-vectors and x-vectors on three different evaluation sets. In the case of i-vectors, discriminative training worked better than generative training. In the case of x-vectors, only generative training was tried to date, and this gave record performance on SRE'16 and SITW.

In future work, we plan to try discriminative training for the HT-PLDA backend also on x-vectors. After that, we want to backpropagate the discriminative training, \emph{through} this backend and also into the x-vector extractor. The idea is that the variable precisions of HT-PLDA should serve as a vehicle for uncertainty propagation from the input MFCCs to the output scores, as more fully motivated in~\cite{Brummer_Odyssey18,meta_embeddings}.

\section{Acknowledgements}
This work was started at the Johns Hopkins University HLTCOE SCALE 2017 Workshop. The authors thank the workshop organizers for inviting us to attend and (in the case of Niko Br\"ummer) for generous travel funding. The work was also supported by Czech Ministry of Interior project No. VI20152020025 ``DRAPAK'' Google Faculty Research Award program,  Technology Agency of the Czech Republic project No. TJ01000208 ``NOSICI'', and by Czech Ministry of Education, Youth and Sports from the National Programme of Sustainability (NPU II) project ``IT4Innovations excellence in science - LQ1602''.

\bibliographystyle{IEEEtran}
 \bibliography{vb4htplda_IS18}

\end{document}